\documentclass[conference]{IEEEtran}
\IEEEoverridecommandlockouts
\usepackage{cite}
\usepackage{amsmath,amssymb,amsfonts}
\usepackage{algorithmic}
\usepackage{graphicx}
\usepackage{textcomp}
\usepackage{xcolor}
\usepackage[utf8]{inputenc}
\usepackage{subcaption}
\usepackage{booktabs, multirow}       
\usepackage{hyperref}

\newcommand\norm[1]{\left\lVert#1\right\rVert}

\def\BibTeX{{\rm B\kern-.05em{\sc i\kern-.025em b}\kern-.08em
    T\kern-.1667em\lower.7ex\hbox{E}\kern-.125emX}}
\begin{document}

\title{Aligning Bird-Eye View Representation of Point Cloud Sequences using Scene Flow \\
}

\author{\IEEEauthorblockN{Minh-Quan Dao, Vincent Frémont, Elwan Héry}
\IEEEauthorblockA{\textit{École Centrale de Nantes} \\
\textit{Laboratoire des Sciences du Numérique de Nantes (LS2N)}\\
Nantes, France \\
firstname.lastname@ec-nantes.fr}
}

\maketitle

\begin{abstract}
Low-resolution point clouds are challenging for object detection methods due to their sparsity. Densifying the present point cloud by concatenating it with its predecessors is a popular solution to this challenge. Such concatenation is possible thanks to the removal of ego vehicle motion using its odometry. This method is called Ego Motion Compensation (EMC). Thanks to the added points, EMC significantly improves the performance of single-frame detectors. However, it suffers from the shadow effect that manifests in dynamic objects' points scattering along their trajectories. This effect results in a misalignment between feature maps and objects' locations, thus limiting performance improvement to stationary and slow-moving objects only. Scene flow allows aligning point clouds in 3D space, thus naturally resolving the misalignment in feature spaces. By observing that scene flow computation shares several components with 3D object detection pipelines, we develop a plug-in module that enables single-frame detectors to compute scene flow to rectify their Bird-Eye View representation. Experiments on the NuScenes dataset show that our module leads to a significant increase (up to 16\%) in the Average Precision of large vehicles, which interestingly demonstrates the most severe shadow effect. The code is published at \url{https://github.com/quan-dao/pc-corrector}.
\end{abstract}


\section{Introduction}
Object detection is a fundamental module of any autonomous driving software stack. While tremendous advancement has been made in camera-based 3D object detection, LiDAR-based methods still dominate public benchmarks thanks to the accurate depth of point clouds. The accuracy of LiDAR-based object detection models essentially correlates to the number of points on each object, which depends on a LiDAR's resolution and distance from objects to the LiDAR. Since these two factors are extrinsic to single-frame methods, their performances are inevitably capped.

Methods operating on point cloud sequences (i.e., multi-frame models) are an appealing alternative. Besides temporal information, a point cloud sequence offers a higher number of points compared to individual sweeps, thus increasing the coverage of objects, especially those at distances. The challenge in developing multi-frame detectors is devising the optimal use of point cloud sequences. 

A popular approach, called Ego Motion Compensation (EMC), is to concatenate a sequence of point clouds in the ego vehicle frame in the present (i.e., the most recent time step). EMC removes ego-motion by transforming past sweeps to the ego vehicle frame in the present using its odometry. First introduced in \cite{caesar2020nuscenes}, EMC has become the standard for object detection on low-resolution point clouds \cite{zhu2019class, yang20203dssd, yin2021center, wang2021pointaugmenting, bai2022transfusion}. The best advantage of EMC is that it enables single-frame methods to enjoy a performance boost thanks to denser point clouds without changing their architecture. It is worth noticing that using EMC on any single-frame method effectively converts it to a multi-frame one.

\begin{figure*}[htb]
    \centering
    \captionsetup[subfigure]{labelformat=empty}
    \begin{minipage}{0.65\linewidth}
        \begin{subfigure}[b]{.45\linewidth}
            \centering
            \includegraphics[width=0.75\textwidth, height=4.05cm]{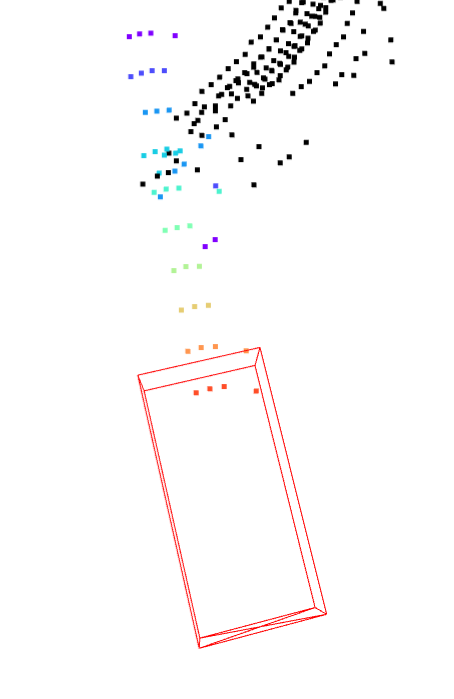}  
            \caption{(a) Object's points by EMC}
            \label{fig:emc-obj}
        \end{subfigure}
        \begin{subfigure}[b]{.45\linewidth}
            \centering
            \includegraphics[width=0.75\textwidth, height=4.05cm]{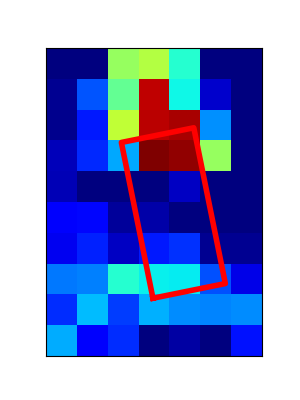}
            \caption{(b) EMC BEV representation}
            \label{fig:ori-bev}
        \end{subfigure}

        \begin{subfigure}{.45\linewidth}
            \centering
            \includegraphics[width=0.75\textwidth, height=4.05cm]{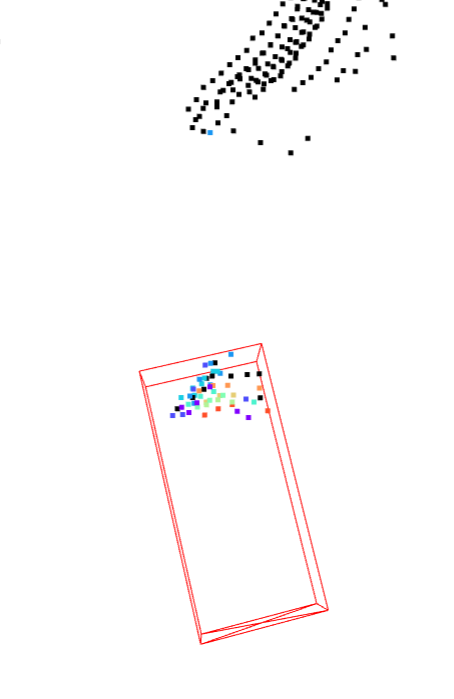}
            \caption{(c) Rectified object's points}
            \label{fig:corrected-obj}
        \end{subfigure}
        \begin{subfigure}{.45\linewidth}
            \centering
            \includegraphics[width=0.75\textwidth, height=4.05cm]{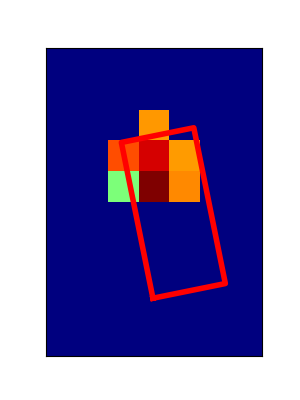}
            \caption{(d) Rectified BEV representation}
            \label{fig:rect-bev}
        \end{subfigure}
    \end{minipage}
    \begin{minipage}{0.3\linewidth}
        \begin{subfigure}{\linewidth}
            \centering
            \includegraphics[width=\textwidth]{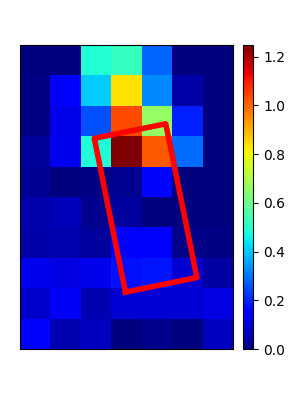}
            \caption{(e) Fused BEV representation}
            \label{fig:fused-bev}
        \end{subfigure}
    \end{minipage}
    \caption{Comparison between BEV representation of a dynamic object before and after rectification. The ground truth object is denoted by the red cuboid in 3D and red rectangle in the BEV plane. Foreground points are color coded such that the hotter their color, the more recent they are.}
\end{figure*}

The major drawback of EMC is the shadow effect \cite{luo2018fast} that manifests in dynamic objects' points scattering along their trajectories (Fig.~\ref{fig:emc-obj}). This misalignment in 3D space results in a misalignment in the feature space, shown in Fig.~\ref{fig:ori-bev}, which limits the performance gain brought by adding past point clouds using EMC to stationary and slow-moving objects only \cite{yang20213d}. As a result, we seek to improve the performance of EMC-boosted single-frame methods by resolving the feature misalignment.

Prior object detection methods that explicitly address the feature misalignment issue can be divided into two categories that align either (i) BEV representation or (ii) object proposals' features. The former evolves from concatenating BEV feature maps \cite{luo2018fast} to sequentially mapping Range-view representation from one time step to another using a warp function made of the rigid transformation \cite{laddha2021mvfusenet}. Its current state is using temporal layers such as Long Short-Term Memory (LSTM) \cite{huang2020lstm} to fuse multi-frame features. 3D-MAN \cite{yang20213d} is an exemplar of the latter category. It generates object proposals independently for each point cloud and stores them in a memory bank. Features of proposals at the target time step get refined by querying the memory bank.

A shortcoming of the methods mentioned above is the lack of explicit supervision of the alignment operation because the notion of "\textit{well-aligned}" is challenging to establish in the feature space. On the other hand, how well two point clouds align in 3D space can be straightforwardly measured using scene flow metrics. For this reason, we devise our feature alignment strategy based on rectifying EMC-concatenated point clouds using scene flow. In details,

\begin{enumerate}
    \item Points in an EMC-concatenated point cloud are rectified according to their scene flow (Fig.~\ref{fig:corrected-obj}).
    \item The rectified point cloud is used to scatter points' features to the BEV plane to make a rectified BEV representation (Fig.~\ref{fig:rect-bev}).
    \item The rectified BEV representation is fused with the BEV representation of the EMC-concatenated point cloud to obtain the fused BEV representation where feature misalignment is resolved (Fig.~\ref{fig:fused-bev}).
\end{enumerate}

In this paper, we make the following contributions:
\begin{itemize}
    \item We develop a plug-in module that enables single-frame object detection methods to rectify their BEV representations of EMC-concatenated point clouds using scene flow. 
    \item We conduct experiments on the NuScenes dataset, where it is standard to use EMC on single-frame methods due to its low-resolution LiDAR (32 beams). Adding our feature alignment method to PointPillars results in an improvement of up to 16\% in Average Precision (AP). Despite being a multi-task method, our estimation of scene flow on the NuScenes dataset reaches 0.506 average End-Point Error (EPE), which is on par with strong scene flow baselines. 
\end{itemize}


\section{Related Works} \label{sec:related-works}

The pioneering work \cite{luo2018fast} takes the mid-fusion approach to resolving the feature misalignment by processing the concatenation of BEV feature maps with a Convolutional Neural Network (CNN). On the other hand, its following works \cite{casas2018intentnet, liang2020pnpnet, djuric2021multixnet} take the early-fusion approach by concatenating voxelized point clouds, $\left(\frac{H}{\Delta H}, \frac{W}{\Delta W}, \frac{L}{\Delta L}\right)$, along the height dimension, then feeding the result to a CNN. Here,  L and W denote the longitudinal and traversal direction, respectively. Their only difference compared to EMC is that the concatenation occurs after the voxelization instead of before. Moreover, the absence of modules dedicated to feature alignment in their architectures raises the question of how effectively the shadow effect is handled. MVFuseNet \cite{laddha2021mvfusenet}, a more recent work of this category, performs the alignment sequentially by mapping the Range-View representation at each time step to its successor using a warp function based on ego-motion, then refining the result with a CNN. Reference \cite{huang2020lstm} takes a similar approach by fusing features of two consecutive point clouds using the so-called "3D sparse conv LSTM".

3D-MAN \cite{yang20213d} solves the feature misalignment issue using object proposals. It first generates object proposals and their associated features independently for each point cloud and stores them in a memory bank. Then, object proposals in the target point cloud (e.g., the most recent one) refine their features by querying the memory bank via the cross-attention mechanism \cite{vaswani2017attention}. While showing strong performance, \cite{yang20213d} relies on a single-frame detector for per-frame proposals generation, thus requiring sufficiently dense point clouds.

Taking a different approach, we strive to align point clouds of a sequence in 3D space using computing scene flow of dynamic points, which naturally results in well-aligned feature maps. The motivation of our alignment strategy has two folds:
\begin{itemize}
    \item Scene flow pipelines require points' features which can be computed by first converting input point clouds to BEV representation, then interpolating using points' projection on the BEV plane. The BEV representation is also needed for object detections, thus the possibility of combining them.
    \item Using scene flow enables explicitly supervising the feature alignment with a physically meaningful signal.   
\end{itemize}

Our method for computation of scene flow takes inspiration from \cite{huang2022dynamic}. We share their method for obtaining point features by interpolating from the BEV representation of point clouds. However, we exploit the availability of ego vehicle localization to concatenate input point clouds before generating the BEV representation. As a result, we obtain the BEV representation for the entire sequence in one shot.

Since motion only makes sense in the context of objects, many motion estimation and segmentation methods \cite{gojcic2021weakly, huang2022dynamic, chen2021moving, chen2022automatic} require instance segmentation by computationally expensive clustering. Therefore, we predict per-point flow directly from their features to achieve high inference speed. Since this flow is unconstrained, we risk our prediction containing physically infeasible motions (e.g., objects having different parts undergoing different motions). This risk is reduced by adding a simplified version of the "Object motion modeling" of \cite{huang2022dynamic}, which we refer to as the Object Head, to our architecture. The role of this module is to predict a single rigid transformation for each instance. During training, we use the Object Head to guide the learning of per-point scene flow by forcing a consistency between their prediction (more details in Section.~\ref{sec:point-loss}). The correspondence between points and objects is available during training as ground truth, and the Object Head is deactivated during testing. As a result, instance segmentation is no longer necessary. This critical difference compared to \cite{huang2022dynamic} enables our short runtime shown in Tab.~\ref{tab:perf-flow3d}.

\section{Aligning Point Cloud Sequences}

The first step toward aligning point clouds is removing the effect of the ego vehicle motion on LiDAR measurements using Ego Motion Compensation (EMC).

\subsection{Ego Motion Compensation}

Let $\mathcal{X}^{t} = \left\{\mathbf{p}_i^t = [x, y, z] \text{ } | \text{ } i = 1, \dots, N \right\}$ denote a point cloud of size $N$ collected at time $t$ and having points expressed the ego vehicle frame $\mathcal{E}(t)$. A sequence of point clouds $\mathcal{S} = \left\{ \mathcal{X}^{t-K}, \mathcal{X}^{t-K+1}, \dots, \mathcal{X}^{t} \right\}$ spanning from the previous $K$ steps to the current time $t$ is merged according to EMC by transforming every point in each point cloud to a common world frame $\mathcal{W}$ using the ego vehicle pose $^{\mathcal{W}}\mathbf{T}_{\mathcal{E}(t-k)}$.

\begin{equation}
    ^{\mathcal{W}}\mathbf{p}_i^{t-k} = ^{\mathcal{W}}\mathbf{T}_{\mathcal{E}(t-k)} \cdot \mathbf{p}_i^{t-k}
    \label{eq:def_emc}
\end{equation}
Here, $k \in \{0, \dots, K\}$. Notice that the world frame $\mathcal{W}$ can be the map frame or the ego vehicle frame at any time step (e.g., $\mathcal{E}^t$ as for NuScenes common practice).


\subsection{Rectifying EMC Point Clouds}

Since EMC does not account for objects' motions, points belong to dynamic objects are scattered along their trajectories. Let $\mathcal{O}$ denote an object and $^{\mathcal{W}}\mathbf{T}_{\mathcal{O}(t)}$ represent $\mathcal{O}$'s pose in the world frame $\mathcal{W}$ at time $t$. At time step $t-k$, a laser beam emitted from the LiDAR hitting $\mathcal{O}$ produces a 3D point $\mathbf{p}^{t-k}$. The image of $\mathbf{p}^{t-k}$ computed by EMC \eqref{eq:def_emc} can be rectified by a two-step transformation as following

\begin{equation}
    ^{\mathcal{W}}\mathbf{\hat{p}}^{t-k} = ^{\mathcal{W}}\mathbf{T}_{\mathcal{O}(t)} 
        \cdot ^{\mathcal{O}(t-k)}\mathbf{T}_{\mathcal{W}} 
        \cdot ^{\mathcal{W}}\mathbf{p}^{t-k}
    \label{eq:def_local_tf}
\end{equation}
The first transformation in \eqref{eq:def_local_tf}, $^{\mathcal{O}(t-k)}\mathbf{T}_{\mathcal{W}}$, returns the coordinate of $\mathbf{p}^{t-k}$ in $\mathcal{O}$'s body frame, which is constant under the rigid body assumption. The second transformation maps the point from the body frame to the world frame $\mathcal{W}$ using the object pose at time $t, ^{\mathcal{W}}\mathbf{T}_{\mathcal{O}(t)}$. In the rest of this paper, we refer to $^{\mathcal{W}}\mathbf{T}_{\mathcal{O}(t)} \cdot ^{\mathcal{O}(t-k)}\mathbf{T}_{\mathcal{W}}$ as the \textit{rectification transformation}.



\section{Joint Objects Detection and Motion Estimation}

\begin{figure*}
    \centering
      \includegraphics[width=.65\linewidth]{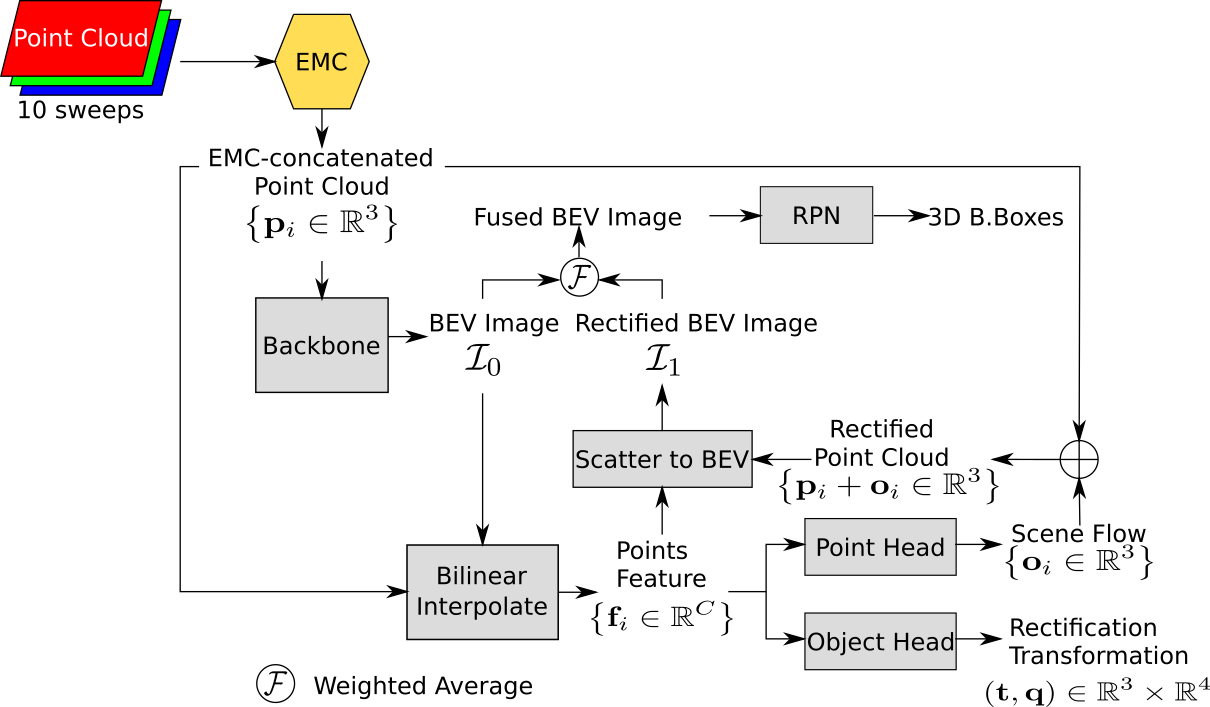}
      \caption{Overview. Our model takes a sequence of point clouds preprocessed by EMC $\left\{\mathbf{p}_i\right\}$ as input. It first voxelizes the point cloud and converts the resulting voxel grid to a BEV image $\mathbf{\mathcal{I}}_0$ using a CNN backbone. Second, points' features $\left\{\mathbf{f}_i\right\}$ are obtained by bilinearly interpolating $\mathbf{\mathcal{I}}_0$ using their projection on the BEV plane. Third, points' features are decoded into objects' rectification transformation and scene flow $\left\{\mathbf{o}_i\right\}$ respectively by Object Head and Point Head. The rectified BEV image $\mathbf{\mathcal{I}}_1$ is the result of scattering $\left\{\mathbf{f}_i\right\}$ back to the BEV using the corrected point cloud $\left\{\mathbf{p}_i + \mathbf{o}_i\right\}$. Then, $\mathbf{\mathcal{I}}_0$ and $\mathbf{\mathcal{I}}_1$ are fused by a weighted sum before being consumed by the RPN to produce 3D bounding boxes.}
      \label{fig:full_pipeline}
\end{figure*}

The pipeline for estimating objects' motion shares several components with object detection models, specifically the computation of the BEV representation, which takes the form of a multi-channel image. Therefore, we propose to combine these two tasks in a unified architecture shown in Fig.~\ref{fig:full_pipeline}.

In our pipeline, an EMC point cloud is voxelized and then processed by the CNN-based backbone to produce a BEV image $\mathbf{\mathcal{I}}_0$. This image is consumed by the Object Motion Estimation branch, made of Point Head and Object Head, to estimate both scene flow and rectification transformation \eqref{eq:def_local_tf}. To demonstrate our method, we choose two backbones: SECOND \cite{yan2018second} and PointPillars \cite{lang2019pointpillars}.

\subsection{Object Motion Estimation}
Points' features are bilinearly interpolated from backbone-made BEV image $\mathbf{\mathcal{I}}_0$ based on their projection to the BEV plane. Given points' features, the Point Head first segments the input point cloud into three classes: background, static foreground, and dynamic foreground. The definition of these classes is as follows:
\begin{itemize}
    \item Background points are those on background objects (e.g., ground, building, traffic lights).
    \item Dynamic foreground points are those on foreground objects that exhibit a translation greater than 0.5 meters during the period of interest (e.g., 0.5 seconds).
    \item Static foreground points are neither background nor dynamic foreground.
\end{itemize}

For a dynamic foreground point $\mathbf{p}^{t-k}$, whose timestamp is $t-k$, the Point Head predicts a scene flow vector $\mathbf{o}^{t-k} \in \mathbb{R}^3$ which is defined as the difference between its location computed by EMC \eqref{eq:def_emc} and by using object trajectory \eqref{eq:def_local_tf}.
\begin{equation}
    \mathbf{o}^{t-k, *} =  ^{\mathcal{W}}\mathbf{\hat{p}}^{t-k} - ^{\mathcal{W}}\mathbf{p}^{t-k}
\end{equation}
Here, $*$ in the superscript denotes the ground truth.

In the Object Head, predicted foreground points are segmented into instances (i.e., objects) which we refer to as \textit{global} groups. Each \textit{global} group is further divided into \textit{local} groups based on foreground points' timestamps.

Let $\mathbf{f}_i \in \mathbb{R}^C$ denote the features of a foreground point $\mathbf{p}_i$. The features $\mathbf{f}_{\mathcal{L}}$ of a \textit{local} group $\mathcal{L}$ is computed as following
\begin{equation}
    \mathbf{f}_{\mathcal{L}} = \text{Max}
        \left(
            \text{cat} 
                \left(
                    \mathbf{f}_i, \text{MLP}\left(\mathbf{p}_i - \mathbf{\Bar{p}}_{\mathcal{L}}\right)
                \right)
                \vert
                \mathbf{p}_i \in \mathcal{L} 
        \right)
\end{equation}
Here, $\mathbf{\Bar{p}}_{\mathcal{L}}$ is the mean coordinate of points in $\mathcal{L}$ and $\text{cat}(\cdot)$ refers to the channel-wise concatenation operation. The features $\mathbf{f}_{\mathcal{G}}$ of a global group $\mathcal{G}$ is the result of max pooling, $\text{Max}\left(\cdot \right)$, of its local groups' features.
\begin{equation}
    \mathbf{f}_{\mathcal{G}} = \text{Max}
        \left(
            \mathbf{f}_{\mathcal{L}_j} 
            \vert 
            \mathcal{L}_j \in \mathcal{G} 
        \right)
\end{equation}

The rectification transformation in \eqref{eq:def_local_tf} is encoded as a 7-vector, which is the concatenation of the translation vector $\mathbf{t} \in \mathbb{R}^3$ and the quaternion $\mathbf{q} \in \mathbb{R}^4$ representing the rotation matrix. This transformation is predicted for each local group $\mathcal{L}_{t-k} (k = 0, \dots, K)$ by an MLP shared among all local groups of all objects.
\begin{equation}
    \left(\mathbf{t}, \mathbf{q}\right)_{\mathcal{L}_{t-k}} = \text{MLP}
        \left[
            \text{cat} 
            \left(
                \mathbf{f}_{\mathcal{L}_{t-k}}, \mathbf{f}_{\mathcal{G}},  \mathbf{\Bar{p}}_{\mathcal{L}_{t-k}}, \mathbf{\Bar{p}}_{\mathcal{L}_{t}}
            \right)
        \right]
\end{equation}

At test time, the input EMC point cloud is rectified by translating dynamic foreground points $\mathbf{p}$ using their scene flow $\mathbf{o}$, instead of local groups' rectification transformation. As a result, the computationally expensive instance segmentation using DBSCAN can be bypassed, thus improving the model's inference speed. 

\subsection{BEV Image Rectification}
The rectified point cloud, obtained by translating points in EMC according to their scene flow, is used to scatter points' features $\left\{\mathbf{f}_i\right\}$ back to the BEV, resulting in the rectified BEV image $\mathbf{\mathcal{I}}_1$. 
Then, $\mathbf{\mathcal{I}}_1$ is fused with backbone-made BEV image $\mathbf{\mathcal{I}}_0$ via a weighted average. 
The weights are computed by stacking two BEV images along the channel dimension, then passing the result to a stack of two 2D Convolution layers with 3-by-3 kernels. 
The rationale of fusing $\mathbf{\mathcal{I}}_1$ with $\mathbf{\mathcal{I}}_0$ is as following. $\mathbf{\mathcal{I}}_1$ is fully sparse because it is created by scattering points’ features in BEV. 
While not possessing the shadow effect, its sparsity harms the detection accuracy \cite{vedder2022sparse}. On the other hand, $\mathbf{\mathcal{I}}_0$, which shows the shadow effect, are not sparse thanks to the occupancy leaking caused by regular 2D convolution layers of the backbone. 
The fusion is to reduce the sparsity and prevent the shadow effect.

\subsection{Region Proposal Network}
The fused BEV image is consumed by a Region Proposal Network (RPN) to produce object detections as 3D bounding boxes. We use the anchor-based \cite{yan2018second} and the center-based \cite{yin2021center} RPN for SECOND and PointPillars, respectively. 

The anchor-based RPN places two anchors in two orthogonal directions for each class of objects at each location of the BEV image and estimates the objectness of each anchor. For each positive anchor, the RPN also predicts a refinement vector to make the anchor fit tighter to its ground truth.

On the other hand, the center-based RPN encodes each ground truth object as a Gaussian on the BEV plane. The mean and covariance of each Gaussian are defined by its corresponding object's center and size, respectively. Then, it uses a series of Convolution layers to decode the input BEV image into pixel-wise center probability and box attributes (e.g., size and orientation).

\subsection{Loss Function}

Our model is trained end-to-end with a loss function $L$ made of 3 terms corresponding to the loss of  the two heads of the Object Motion Estimation branch and the RPN.
\begin{equation}
    L = L_{\text{objects}} + L_{\text{points}} + L_{\text{RPN}}
    \label{eq:l_stage1}
\end{equation}

\subsubsection{Object Loss}
Let $\left(\mathbf{t}, \mathbf{q}\right)$ and $\left(\mathbf{t}^{*}, \mathbf{q}^{*}\right)$ respectively be the prediction and ground truth of the rectification transformation of a local group $\mathcal{L}$. The object loss of this local group is 
\begin{equation}
    L_{\text{objects}, \mathcal{L}} = \text{smooth}_{L_1}\left(\mathbf{t} - \mathbf{t}^{*} \right) 
        + \norm{
            \text{Rot}\left(\mathbf{q}\right) - \text{Rot}\left(\mathbf{q}^{*}\right)
        }_{F}
        + L_{\text{recon}}
    \label{eq:l_local_individual}
\end{equation}
Here, $\norm{\cdot}_{F}$ denotes the Frobenius norm. $\text{Rot}\left(\cdot\right)$ represents the function that converts a quaternion to a rotation matrix. $L_{\text{recon}}$ is the difference between points of $\mathcal{L}$ transformed by the prediction and ground truth of the rectification transformation
\begin{equation}
    L_{\text{recon}} = \frac{1}{N_{\mathbf{p}}^{\mathcal{L}}} \sum_{\mathbf{p} \in \mathcal{L}} \text{smooth}_{L_1} \left(
        \text{T}\left(\mathbf{t}, \mathbf{q}\right) \cdot \mathbf{p} 
        - \text{T}\left(\mathbf{t}^{*}, \mathbf{q}^{*}\right) \cdot \mathbf{p} 
    \right)
    \label{eq:l_recon}
\end{equation}
In \eqref{eq:l_recon}, $N_{\mathbf{p}}^{\mathcal{L}}$ is the number of points in the local group $\mathcal{L}$. $\text{T}(\mathbf{t}, \mathbf{q})$ denotes the function that converts a translation vector $\mathbf{t}$ and a quaternion $\mathbf{q}$ to a homogenous transformation matrix.

$L_{\text{objects}}$ in \eqref{eq:l_stage1} is the sum of applying \eqref{eq:l_local_individual} to every local group $\mathcal{L}$ of every global group $\mathcal{G}$
\begin{equation}
    L_{\text{objects}} = \frac{1}{N_{\mathcal{L}}} \sum_{\mathcal{G}} \sum_{\mathcal{L} \in \mathcal{G}} L_{\text{objects}, \mathcal{L}}
\end{equation}
where, $N_{\mathcal{L}}$ is the total number of local groups.

\subsubsection{Point Loss} \label{sec:point-loss}
The loss of Object Motion Estimation's Point Head is made of classification loss, offset loss and consistent loss.
\begin{equation}
    L_{\text{points}} = L_{\text{cls}} + L_{\text{offset}} + L_{\text{consistent}}
\end{equation}

Following \cite{huang2022dynamic}, we use the sum of weighted binary cross entropy loss $\text{L}_{\text{bce}}$ and Lovasz-Softmax loss $\text{L}_{\text{ls}}$ \cite{berman2018lovasz} as the classification loss $L_{\text{cls}}$.
\begin{equation}
    L_{\text{cls}} = 
        \frac{1}{N_{\mathbf{p}}} \sum_{\mathbf{p}}
         \text{L}_{\text{bce}} \left(\mathbf{c}, \mathbf{c}^{*}\right) 
        + \text{L}_{\text{ls}} \left(\mathbf{c}, \mathbf{c}^{*}\right)
    \label{eq:l_point_cls}
\end{equation}
In \eqref{eq:l_point_cls}, $\mathbf{c} \text{ and } \mathbf{c}^{*}$ are the prediction and ground truth of the class probability of a point $\mathbf{p}$. $N_{\mathbf{p}}$ is the number of points in the inputted EMC point cloud.

$L_{\text{offset}}$ measures the difference between dynamic foreground points $\mathbf{p}_{+}$ translated by predicted offset vectors $\mathbf{o}$ and their position after undergone the ground truth rectification transformation $\left(\mathbf{t}^{*}, \mathbf{q}^{*}\right)$.
\begin{equation}
    L_{\text{offset}} = \frac{1}{N_{\mathbf{p}_{+}}} \text{smooth}_{L_1} \left(
        \mathbf{p}_{+} + \mathbf{o} - \text{T}\left(\mathbf{t}^{*}, \mathbf{q}^{*}\right) \cdot \mathbf{p}_{+}
    \right)
\end{equation}
$N_{\mathbf{p}_{+}}$ is the number of dynamic foreground points in the input point cloud.

The Object Head groups points to local groups before predicting a rectification transformation for each group, which is then applied to every point inside a group. For this reason, it enforces the rigid motion among dynamic objects which is a realistic assumption in the context of autonomous driving. On the other hand, the Point Head predicts an unconstrained offset vector for each dynamic point, thus risking rectified point clouds containing physically infeasible objects (e.g., deformed cars due to different parts undergoing different transformations). We reduce this risk by enforcing the consistency between predictions made by the two heads using $L_{\text{consistent}}$. 
\begin{equation}
    L_{\text{consistent}} = \frac{1}{N_{\mathbf{p}_{+}}} \text{smooth}_{L_1} \left(
        \mathbf{p}_{+} + \mathbf{o} - \text{T}\left(\mathbf{t}, \mathbf{q}\right) \cdot \mathbf{p}_{+}
    \right)
\end{equation}

\subsubsection{RPN}
The loss function $L_{\text{RPN}}$ of the anchor-based and center-based RPN are respectively defined in \cite{yan2018second} and \cite{zhou2019objects}.


\section{Experiments}


\subsection{Dataset and Evaluation Setting}

\subsubsection{NuScenes}
The NuScenes dataset \cite{caesar2020nuscenes} contains 700 scenes for training and 150 scenes for validation. Each comprises data collected by a multimodal sensor suite of an autonomous vehicle over approximately 20 seconds. The sensor suite has one 32-beam LiDAR that operates at 20 Hz. Once all sensors are in sync, a keyframe is established. The frequency of sensor synchronization is 2 Hz. Objects' ID and location, in the form of bounding boxes, are annotated for every keyframe.

\subsubsection{Metrics}
The object detection task is primarily evaluated by the Average Precision (AP) implemented by NuScenes, which defines a match based on the 2D distance on the BEV plane instead of the intersection over union (IoU). To have a more complete evaluation that also takes bounding boxes' size and orientation, we use AP calculated by matching prediction and ground truth using 3D IoU as a secondary metric. Following prior works \cite{djuric2021multixnet, laddha2021mvfusenet}, we set the IoU matching threshold to 0.7, 0.1, and 0.3 for cars, pedestrians, and bicyclists. Since the detection of other classes is not addressed in prior works, we set the threshold for trucks, construction vehicles, buses, trailers, barriers, motorcycles, and traffic cones to 0.7, 0.7, 0.7, 0.7, 0.5, 0.5, 0.5, respectively.   

The Object Motion Estimation is evaluated by standard scene flow metrics \cite{liu2019flownet3d} including 3D End-Point Error (EPE), strict/ relaxed accuracy (AccS/ AccR), and outlier (ROutliers). The EPE is the mean of the 3D distance between rectified points and their ground truth. The AccS/ AccR is the percentage of points having either EPE $<$ 0.05/ 0.10 meters or relative error $<$ 0.05/ 0.10. The ROutliers is the percentage of points with EPE $>$ 0.30 meters and relative error $>$ 0.30.


\subsection{Implementation Details}

We follow the common approach to handle the sparsity of NuScenes point clouds that concatenating a keyframe point cloud with all non keyframe point clouds between itself and its predecessor using EMC \cite{caesar2020nuscenes, zhu2019class, yang20203dssd, yin2021center, wang2021pointaugmenting, bai2022transfusion}. Let $t$ denote the time step of the keyframe. The world frame $\mathcal{W}$ is set at the LiDAR frame at time step $t$. The point cloud of a non keyframe collected at timestamp $t - k \text{ } (k = 1, \dots, 9)$ are mapped from the LiDAR frame at this time step to the world frame $\mathcal{W}$ using ego vehicle poses and LiDAR calibration.

The ground truth of rectification transformation in \eqref{eq:def_local_tf} requires knowing objects' poses in the world frame $\mathcal{W}$ at the keyframe timestamp $^{\mathcal{W}}\mathbf{T}_{\mathcal{O}(t)}$ and non keyframe timestamp $^{\mathcal{W}}\mathbf{T}_{\mathcal{O}(t-k)}$. Since NuScenes only provides objects' poses in keyframes, we obtain their poses in non keyframes by linearly interpolating annotations of two keyframes that are respectively prior and successor to each non keyframe.

\begin{figure*}[htb]
    \centering
    \begin{subfigure}{.47\linewidth}
      \centering
      \includegraphics[width=.9\linewidth]{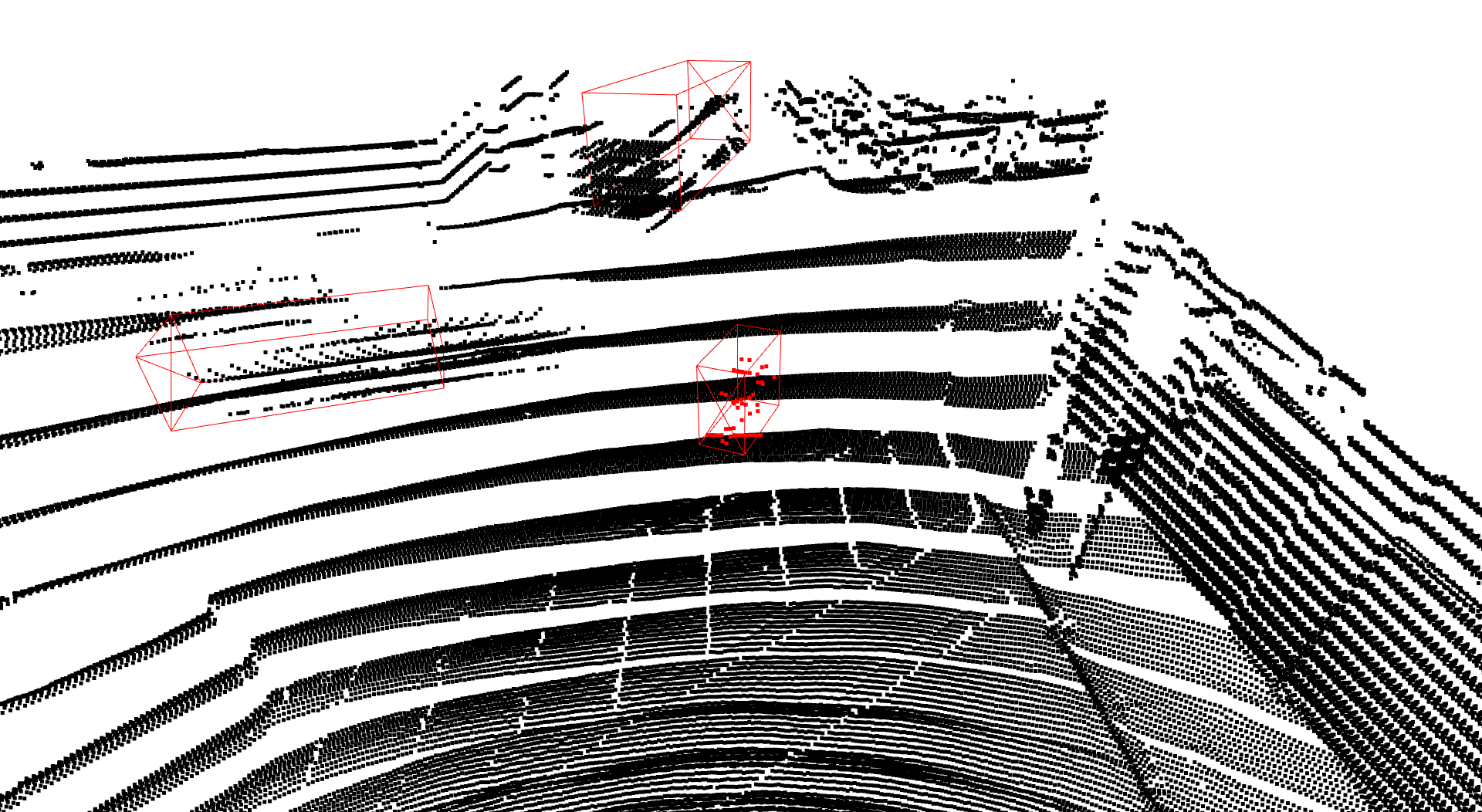}
      \caption{Ground truth sampling}
    \end{subfigure}%
    \begin{subfigure}{.47\linewidth}
      \centering
      \includegraphics[width=.9\linewidth]{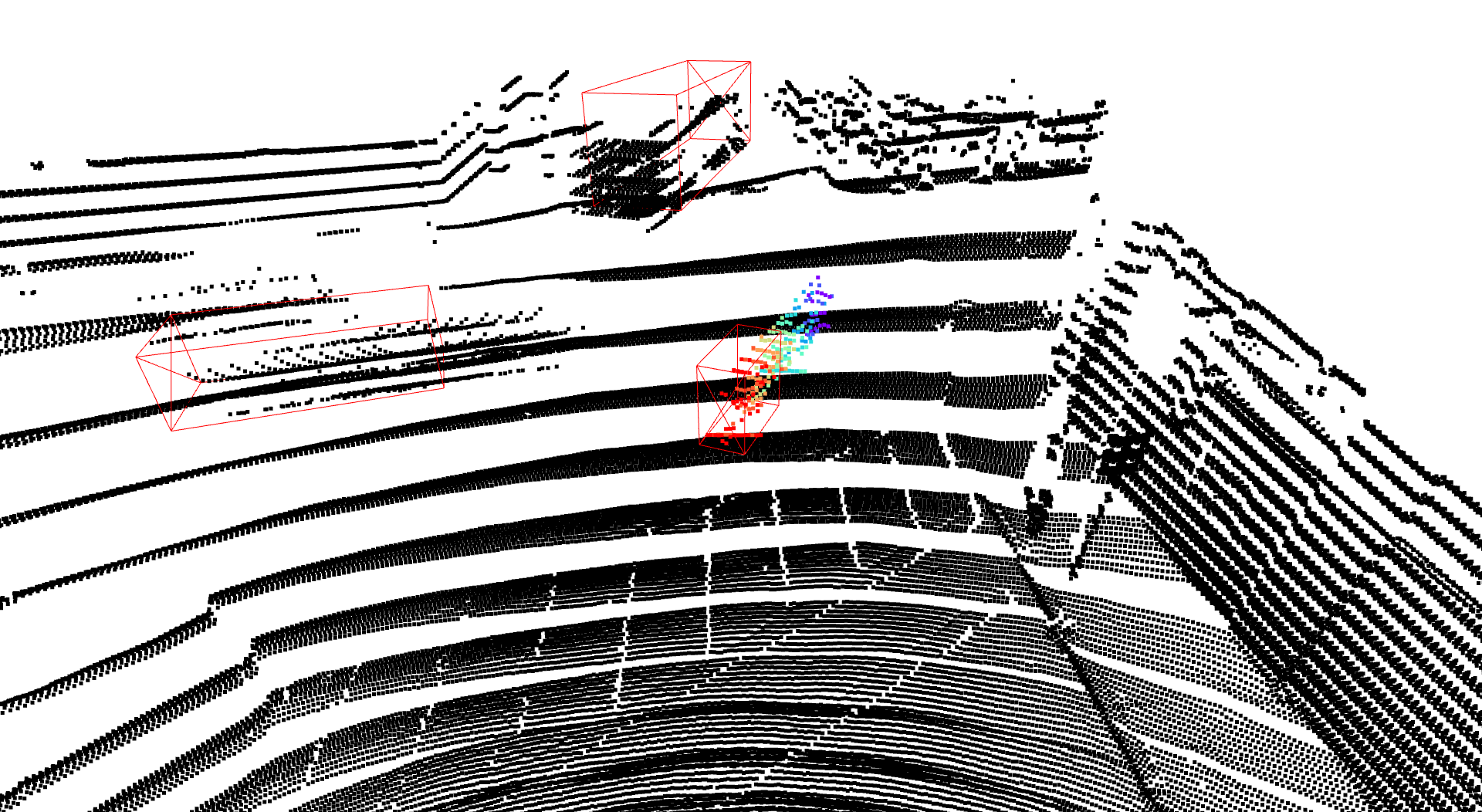}
      \caption{Ground truth sampling and points on object's trajectory (ours)}
    \end{subfigure}

    \caption{The comparison between scene augmented by the ground truth sampling strategy and by our strategy. Points belong to the added object are colored coded according to their timestamp. The hotter the color, the more recent timestamp.}
    \label{fig:comparison_gt_sampling_traj_sampling}
\end{figure*}

To improve the generalization of our model, the following geometric transformations are applied to point clouds and ground truth: random flip along the x- and y-axis of the world frame $\mathcal{W}$, global scaling with a factor sampled from $\mathcal{U}_{[0.95, 1.05]}$, and global rotation by an angle sampled from $\mathcal{U}_{[-\pi / 8, \pi / 8]}$ around the z-axis of the world frame $\mathcal{W}$. Here, $\mathcal{U}$ denotes the uniform distribution. Furthermore, we adopt the ground truth sampling strategy of \cite{zhu2019class} which randomly takes boxes and their points from a database and places them in the input point cloud. Notably, we introduce a modification to this sampling strategy by complementing each sampled ground truth with points in its trajectory spanning from the current time step to 10 steps in the past. This modification is similar to the "Sequence GtAug" of \cite{xu2022int}. Its motivation is to increase the number of moving objects in each point cloud, thus increasing the amount of supervision on the Object Motion Estimation. A comparison between the regular ground truth sampling and our modified version is shown in Fig.~\ref{fig:comparison_gt_sampling_traj_sampling}.

In our experiments, input point clouds are limited to the range of $[-51.2, 51.2] \times [-51.2, 51.2] \times [-5.0, 3.0]$ meters along X-, Y-, and Z-axis of the world frame $\mathcal{W}$ and the voxel size is set to $(0.1, 0.1, 0.2)$. We use OpenPCDet \cite{openpcdet2020} for our implementation. Further details on the model's hyperparameters can be found in our code release. 

Due to the large size of the NuScenes dataset, we only train our model on a quarter of the training set. This mini partition of the training set is obtained by sorting keyframes by their point clouds' timestamp, then taking one every four keyframes. Our model is trained for 25 epochs with a total batch size of 16 distributed over 8 GPUs with sync batch norm. The optimizer is set to AdamW \cite{loshchilov2018fixing}. The learning rate is regulated by the One Cycle scheduler \cite{smith2018disciplined} with the maximum value of $0.003$ for SECOND and $0.001$ for PointPillars. It is worth noticing that the evaluation takes place on the entire validation set.

\subsection{Results}
The comparison of our model against methods specialized in estimating scene flow is shown in Tab.~\ref{tab:perf-flow3d}. Since we only predict scene flow for dynamic foreground points, the comparison in Tab.~\ref{tab:perf-flow3d} only accounts for these points. More importantly, we evaluate scene flow predicted by the Point Head of the Object Motion Estimation branch because the Object Head is deactivated at test time to avoid the computationally expensive instance segmentation using DBSCAN. In Tab.~\ref{tab:perf-flow3d} and following tables, the best and second-best performances are marked by \textbf{bold} and \underline{underscore} font, respectively

While being trained on fewer data (a quarter of NuScenes training set) and not having an architecture optimized solely for scene flow, our model outperforms PPWC-Net and FLOT and is on par with WsRSF and NSFPrior. Notably, our model is the second best in the metric EPE.

In addition, we report the runtime of our models measured in seconds on an NVIDIA A6000 GPU. The five baselines presented in Tab.~\ref{tab:perf-flow3d} have their runtime measured on an NVIDIA RTX 3090 GPU (as reported by \cite{huang2022dynamic}), which has a similar computing capability. Compared to them, our models achieve significantly better runtimes.

To verify the impact of aligning BEV representation using scene flow on the object detection performance, we modify SECOND and PointPillars to resemble the architecture shown in Fig.~\ref{fig:full_pipeline} and train them end-to-end. Tab.~\ref{tab:obj-det-performance} shows an improvement brought by our module for most classes. Notably, trucks, construction vehicles, and buses, which exhibit severe shadow effects during motion due to their large size, enjoy significant performance gain (up to 7.8 AP or 16\%). Furthermore, the detection improvement is higher on PointPillars since its BEV images have double the size of SECOND's, thus more severe feature misalignment. This highlights the importance of handling the misalignment between features and objects' locations. Interestingly, pedestrians also experience 0.8 AP improvement even though their motions violate the rigid body assumption made in \eqref{eq:def_local_tf}. 

\begin{table}[tb]
    \caption{Performance of our model on scene flow metrics}
    \label{tab:perf-flow3d}
    \centering
    \resizebox{\linewidth}{!}
    {
    \begin{tabular}{l c c c c c}
    \toprule
    Method & EPE $\downarrow$ & AccS $\uparrow$ & AccR $\uparrow$ & ROutliers $\downarrow$ & Runtime $\downarrow$ \\
    \midrule
    FLOT \cite{puy2020flot} & 1.216 & 3.0 & 10.3 & 63.9 & 2.01 \\
    
    NSFPrior \cite{li2021neural} & 0.707 & \underline{19.3} & \underline{37.8} & 32.0 & 63.46 \\
    
    PPWC-Net \cite{wu2020pointpwc} & 0.661 & 7.6 & 24.2 & 31.9 & 0.99 \\
    
    WsRSF \cite{gojcic2021weakly} & 0.539 & 17.9 & 37.4 & \underline{22.9} & 1.46 \\

    PCAccumulation \cite{huang2022dynamic} & \textbf{0.301} & \textbf{26.6} & \textbf{53.4} & \textbf{12.1} & 0.25\\
    \midrule
    Our PointPillars & 0.547 & 14.5 & 26.2 & 36.9 & \textbf{0.06} \\

    Our SECOND & \underline{0.506} & 16.8 & 30.2 & 33.8 & \underline{0.09} \\

    \bottomrule
    \end{tabular}
    }
\end{table}

\begin{table*}
    \caption{Object detection results on NuScenes dataset evaluated by matching based on distance on BEV/ IoU. All models are trained on a quarter of the training set and evaluated on the entire validation set}
    \label{tab:obj-det-performance}
    \centering
    \resizebox{\linewidth}{!}
    {
    \begin{tabular}{c c c c c c c c c c c c c c}
      \toprule
           & Car & Truck & Const. & Bus & Trailer & Barrier & Motor. & Bicyc. & Pedes. & Traff. & mAP \\
      \midrule
      SECOND & 
        73.8/ 32.8 & 28.5/ 14.6 & 12.4/ 2.1 & 43.7/ 22.6 & 32.2/ 12.2 & 48.3/ 4.5 & 21.0/ 20.3 & 5.0/ 11.2 & 69.5/ 61.8 & 40.4/ 4.2 & 37.5/ 18.6 \\
      + our module & 
        \textbf{+0.8}/ \textbf{+1.1}  & \textbf{+2.8}/ \textbf{+0.6} & \textbf{+1.6}/ \textbf{+0.4} & \textbf{+5.1}/ \textbf{0.6} & -1.2/ -2.0 & \textbf{+2.1}/ \textbf{+1.0} & \textbf{+3.0}/ \textbf{+2.2} & -0.2/ -0.3 & \textbf{+0.8}/ \textbf{+1.0} & \textbf{+0.6} \textbf{+0.3} & \textbf{+1.5}/ \textbf{+0.5} \\
      \midrule
      PointPillars & 78.9/ 27.0 & 37.9/ 13.2 & 4.2/ 0.3 & 48.9/ 20.8 & 21.4/ 4.0 & 48.4/ 3.8 & 28.1/24.9 & 7.3/ 15.8 & 73.3/ 65.4 & 41.5/ 2.5 & 39.0/ 17.8 \\
      + our module & \textbf{+1.8}/ \textbf{+5.0} & \textbf{+7.3}/ \textbf{+3.7} & \textbf{+2.7}/ -0.2 & \textbf{+7.8}/ \textbf{+6.1} & \textbf{+6.3}/ \textbf{+2.7} & -1.2/ \textbf{+0.3} & \textbf{+5.1}/ \textbf{+3.1} & \textbf{+0.3}/ -0.7 & \textbf{+0.8}/ \textbf{+0.7} & \textbf{+3.5}/ \textbf{+1.0} & \textbf{+3.4}/ \textbf{+2.2} \\
      \bottomrule
    \end{tabular}
    }
\end{table*}


The comparison of our models against other multi-frame methods on NuScenes is shown in Tab.~\ref{tab:obj-det-performance-iou}. Here, the matching between predictions and ground truth is based on the IoU. Our models exhibit strong performance in the class Pedestrians. As can be seen, our PointPillars $\frac{1}{2}$ exceeds the 66.1 AP of MultiXNet by 2.7 AP to be the second-best model despite being trained on only half of the data used for the baselines. We hypothesize that the removal of the shadow effect on pedestrians helps improve our models' generalization, thus getting high performance from fewer training data.

\begin{table}
    \caption{Object detection results on NuScenes dataset compared to other multi-frame methods. The numbers following models' name denote the fraction of NuScenes training set used for training.}
    \label{tab:obj-det-performance-iou}
    \centering
    \begin{tabular}{l c c c}
      \toprule
        & Pedestrians & Cars & Bicyclists \\
      \midrule
      IntentNet \cite{casas2018intentnet}  & 63.4 & 60.3 & 31.8 \\
      MultiXNet \cite{djuric2021multixnet} & 66.1 & \underline{60.6} & \underline{32.6} \\
      MVFuseNet \cite{laddha2021mvfusenet} & \textbf{76.4} & \textbf{67.8} & \textbf{44.5} \\
      \midrule
      Our SECOND $\frac{1}{4}$ & 62.8 & 33.9 & 10.9 \\
      Our PointPillars $\frac{1}{2}$ & \underline{68.8} & 40.5 & 29.3 \\
      \bottomrule
    \end{tabular}
\end{table}

On the other hand, we explain the gap between our models and baselines in class Cars and Bicyclists by the small-size training set. Due to limited computational resources, we only use up to half of the NuScenes training set to keep our experiments affordable. Tab.~\ref{tab:perf-scale} shows that scaling from one-eight to half of NuScenes training leads to a 17.8 and 19.3 increase in the AP of Cars and Bicyclists. Therefore, we believe our performance can greatly improve if more resources are available. Furthermore, the three baselines employ HDMap as an additional input, which provides a strong inductive bias for estimating cars' orientation and eliminating False Positive detections. Last but not least, our models are trained with a smaller mini-batch size, which is 16 compared to 32 and 64 of MultiXNet and MVFuseNet. As pointed out by \cite{peng2018megdet}, a small mini-batch size can hurt detection performance by (i) failing to provide accurate statistics for the Batch Normalization layer and (ii) possessing an imbalance number of positive and negative examples (e.g., the dominant of negative proposals at the early stage).

\begin{table}[tb]
    \caption{Evolution of the performance of our PointPillars with respect to the portion of the NuScenes training set that is actually used for training.}
    \label{tab:perf-scale}
    \centering
    \begin{tabular}{c c c c c}
      \toprule
       Size of actual & \multirow{2}{*}{Pedestrians} & \multirow{2}{*}{Cars} & \multirow{2}{*}{Bicyclists} & Training Time \\
       training set &                            &                       &                             & (hours) \\
        \midrule
      $1 / 8$   & 59.4 & 22.7 & 10.0 & 15 \\
      $1 / 4$   & 66.1 & 32.0 & 15.1 & 30 \\
      $1 / 2$   & 68.8 & 40.5 & 29.3 & 60 \\   
      \bottomrule
    \end{tabular}
\end{table}

\begin{figure*}
    \captionsetup[subfigure]{labelformat=empty}
    \centering
    \begin{subfigure}{0.45\linewidth}
        \centering
        \includegraphics[width=.4\linewidth]{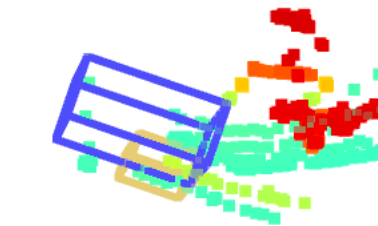}
        \caption{(a)}
    \end{subfigure}%
    \begin{subfigure}{0.45\linewidth}
        \centering
        \includegraphics[width=.4\linewidth]{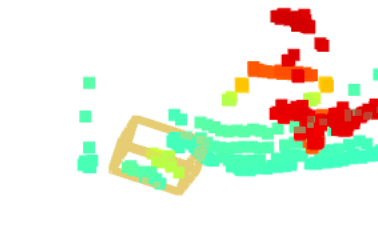}
        \caption{(b)}
    \end{subfigure}%


    \begin{subfigure}{0.45\linewidth}
        \centering
        \includegraphics[width=.42\linewidth]{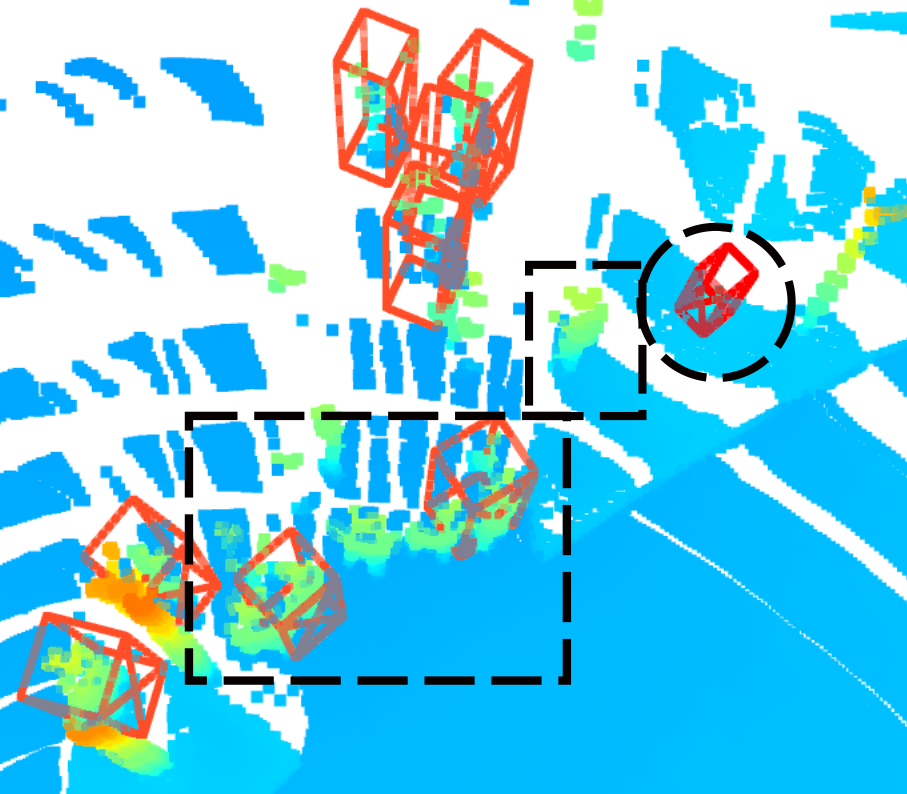}
        \caption{(c) \\ SECOND}
    \end{subfigure}%
    \begin{subfigure}{0.45\linewidth}
        \centering
        \includegraphics[width=.42\linewidth]{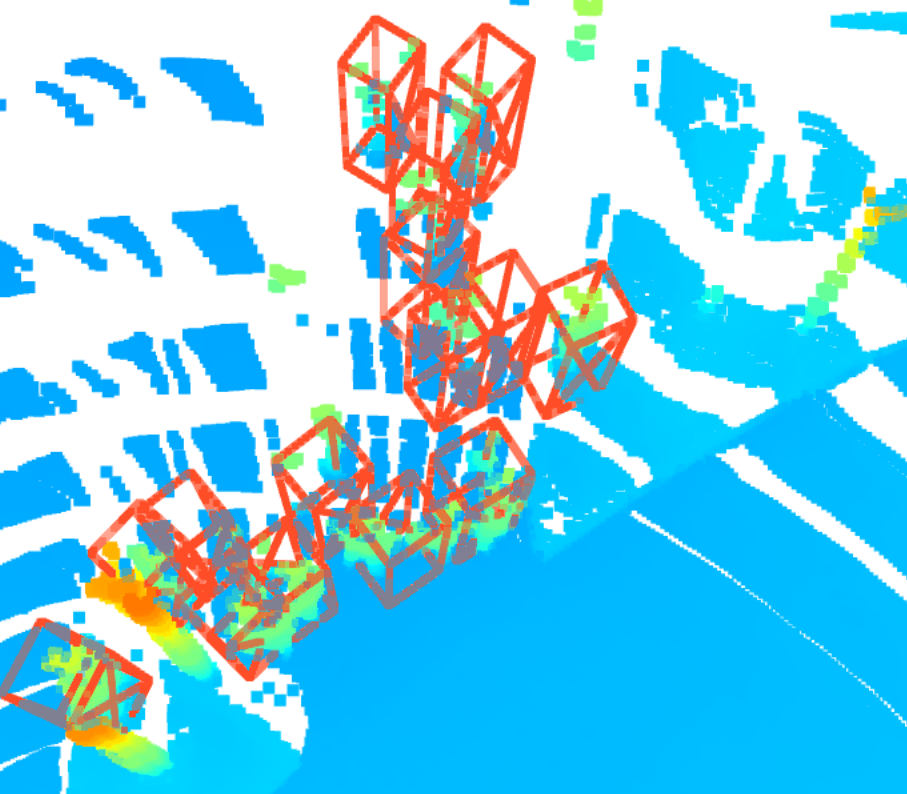}
        \caption{(d) \\ SECOND with BEV representation alignment}
    \end{subfigure}%

    \caption{Qualitative performance on NuScenes}
    \label{fig:qualitative}
\end{figure*}

Fig.~\ref{fig:qualitative} compares the prediction made by the original SECOND and SECOND with BEV representation alignment. Fig.~\ref{fig:qualitative}a shows that SECOND made a false-positive car detection due to the shadow effect of a moving bicyclist. This mistake is avoided as points of the bicyclist are rectified in Fig.~\ref{fig:qualitative}b. The dashed rectangles in Fig.~\ref{fig:qualitative}c mark the region where SECOND made false-negative pedestrian detections. Most false-negative predictions occur when several pedestrians stand in a cluster, thus resembling the scenario where fewer pedestrians walk near each other. In addition, the dashed circle in Fig.4c indicates a false-positive pedestrian detection which is, in fact, a pole. On the other hand, the rectified BEV images help SECOND successfully detect all pedestrians in the cluster and avoid mistaking the pole for a pedestrian (Fig.~\ref{fig:qualitative}d). This observation supports the hypothesis that rectifying the shadow effect improves our models' generalization.

\section*{Acknowledgment}

This work was granted access to the HPC resources of IDRIS under the allocation 2021-AD011012128R1 made by GENCI. This work has been supported in part by the ANR AIby4 under the number ANR-20-THIA-0011. This work was carried out in the framework of the NExT Senior Talent Chair DeepCoSLAM, which were funded by the French Government, through the program Investments for the Future managed by the National Agency for Research (ANR-16-IDEX-0007), and with the support of Région Pays de la Loire and Nantes Métropole.


\bibliography{main.bbl}

\end{document}